\def\year{2022}\relax
\documentclass[letterpaper]{article} 
\usepackage{aaai22}  
\usepackage{times}  
\usepackage{helvet}  
\usepackage{courier}  
\usepackage[hyphens]{url}  
\usepackage{graphicx} 
\usepackage{natbib}  
\usepackage{caption} 
\DeclareCaptionStyle{ruled}{labelfont=normalfont,labelsep=colon,strut=off} 
\usepackage{algorithm}
\usepackage{algorithmic}

\usepackage{newfloat}
\usepackage{listings}

\usepackage{amsmath}
\usepackage{amssymb}
\usepackage{amsfonts}
\usepackage{enumitem}
\usepackage{graphicx}
\usepackage{xcolor}
\graphicspath{{Figure/}}

\floatstyle{ruled}
\newfloat{listing}{tb}{lst}{}
\floatname{listing}{Listing}
\title{Towards Building an Open-Domain Dialogue System \\Incorporated with Internet Memes}
\author{
    Hua Lu\thanks{~First two authors contributed equally to this work.}, 
    Zhen Guo\footnotemark[1],
    Chanjuan Li\thanks{~Work was done during internship at Baidu.}, 
    Yunyi Yang\footnotemark[2], 
    Huang He, 
    Siqi Bao
}
\affiliations{
    Baidu Inc., China\\
    \{luhua05, guozhenguozhen\}@baidu.com
}

\begin{document}

\maketitle

\begin{abstract}
In recent years, Internet memes have been widely used in online chatting. Compared with text-based communication, conversations become more expressive and attractive when Internet memes are incorporated. This paper presents our solutions for the Meme incorporated Open-domain Dialogue (MOD) Challenge of DSTC10, where three tasks are involved: text response modeling, meme retrieval, and meme emotion classification.  Firstly, we leverage a large-scale pre-trained dialogue model for coherent and informative response generation. Secondly, based on interaction-based text-matching, our approach can retrieve appropriate memes with good generalization ability. Thirdly, we propose to model the emotion flow (EF) in conversations and introduce an auxiliary task of emotion description prediction (EDP) to boost the performance of meme emotion classification. Experimental results on the MOD dataset demonstrate that our methods can incorporate Internet memes into dialogue systems effectively. 
\end{abstract}

\section{Introduction}
\begin{figure}[t]
    \centering
    \includegraphics[width=0.48\textwidth]{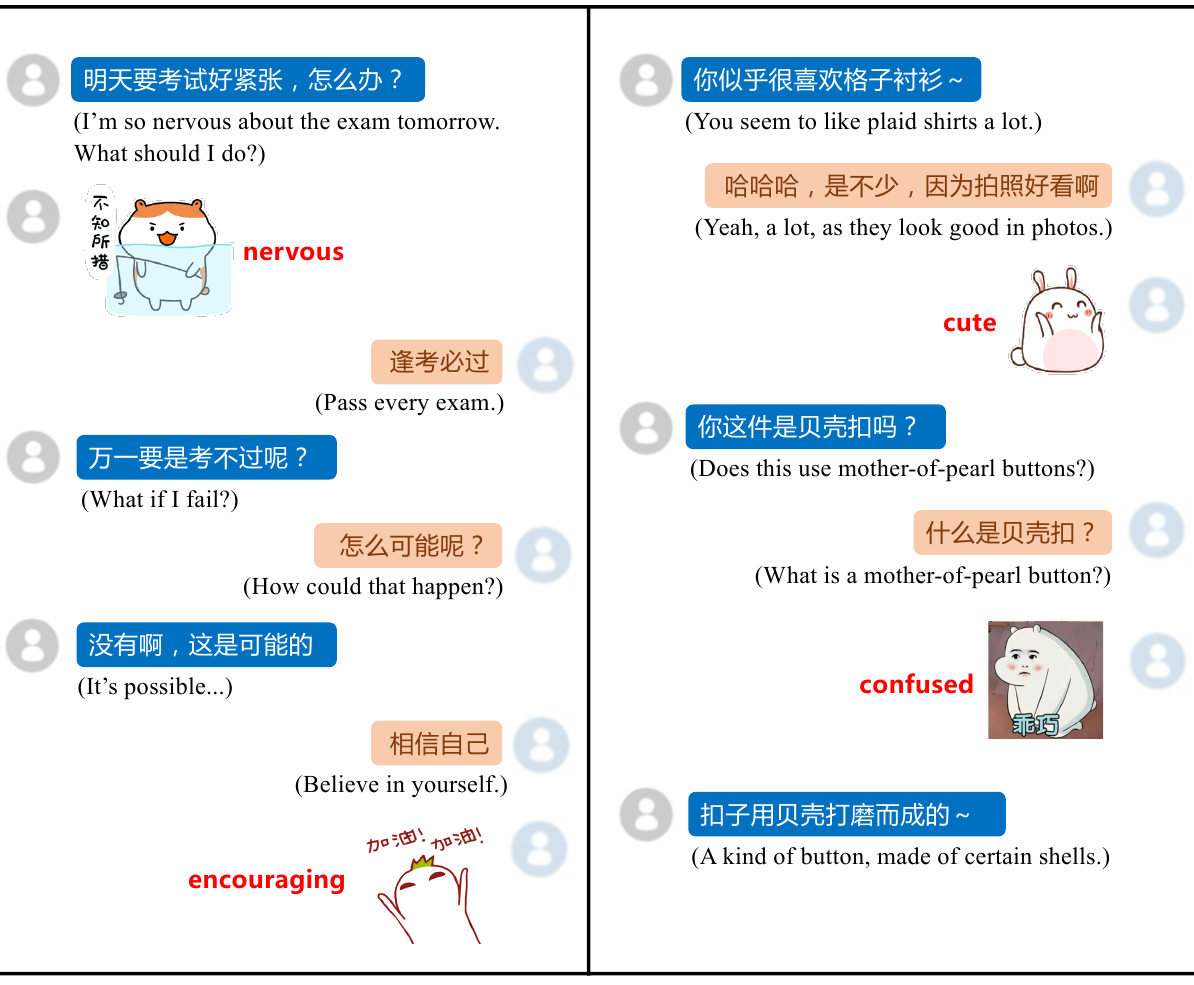}
    \caption{Two examples from the MOD dataset. Corresponding emotion is annotated for each meme in red.}
    \label{fig1:examples}
\end{figure}

As Internet memes can make dialogues more vivid and engaging, nowadays, people tend to incorporate memes when chatting online \cite{kulkarni2017internet,jiang2020exploring}.
Despite that Internet memes have become an effective means of expression, they are rarely considered by most open-domain dialogue systems.
In DSTC10, the Meme incorporated Open-domain Dialogue (MOD) challenge aims to incorporate Internet memes into open-domain dialogues. It includes the following three tasks:
(1) \textbf{Text Response Modeling}: given a multi-modal context, the task here is to generate a coherent and informative text response. (2) \textbf{Meme Retrieval}: given a multi-modal context and a text response, the task aims to retrieve an appropriate meme. (3) \textbf{Meme Emotion Classification}: given a multi-modal context and a text response with a meme, the task here is to predict the emotion type of the Internet meme. Figure \ref{fig1:examples} shows two examples of conversations in the MOD dataset involving texts, Internet memes, and emotions.

In particular, the test set of MOD is divided into an easy test version and a hard test version. The latter, which contains memes not appearing in the train set, is used to evaluate the generalization ability of the dialogue system. 
In this work, we introduce the following solutions for the three tasks:
\begin{itemize}
    \item In Task1, we leverage a powerful pre-trained open-domain dialogue model for coherent and informative text response generation.
    \item In Task2, we represent memes with textual information consisting of meme titles and OCR texts (extracted from the memes). Based on interaction-based text-matching, our approach can retrieve appropriate memes with good generalization ability.
    \item In Task3, we propose to model the emotion flow (EF) in conversations and introduce an auxiliary task of emotion description prediction (EDP) to enhance the ability of meme emotion recognition. 
\end{itemize}
Experimental results demonstrate that our methods can effectively incorporate Internet memes into dialogue systems. Our methods achieve first place in four out of six leaderboards and second place in the others with competitive performance.

\section{Methodology}
Our detailed solutions towards these three tasks will be discussed in the following.

\subsection{Text Response Modeling}
In open-domain conversation, users are free to talk about any topic, and the system's replies are expected to meet a high standard in many aspects, including coherence, consistency, informativeness, etc. Incorporated with Internet memes, the dialogue context can be formulated as $C_t = \{\langle u_1,m_1\rangle, \langle u_2,m_2\rangle,\cdots,\langle u_t,m_t\rangle\}$, where $u_i$ represents the $i$-th utterance text and $m_i$ refers to its associated meme. If no Internet meme is used in the $i$-th utterance, $m_i$ will be denoted as None. The task of text response modeling is to generate the response $r$ (i.e., next utterance $u_{t+1}$) given the multi-modal dialogue context $C_t$. 

As suggested in the MOD baseline \citep{fei2021towards}, the Internet meme can be represented with visual features extracted by EfficientNet \citep{tan2019efficientnet}. While in our preliminary experiments, we observe that incorporating memes, whether as visual features or as textual features, brings little benefit to text response generation. The reasons might be two-fold. Firstly, as memes are usually about emotional expressions with little narrative information, the absence of memes might not remarkably undermine a dialogue system's text response generation ability. Secondly, given that the MOD dataset is collected by inserting memes into existing conversations, the reliance on these memes might be relatively weak for text response generation. Therefore, we treat this task as a standard text-based response generation problem, with memes in the context set as None.

\begin{figure}[t]
    \centering
    \includegraphics[width=0.45\textwidth]{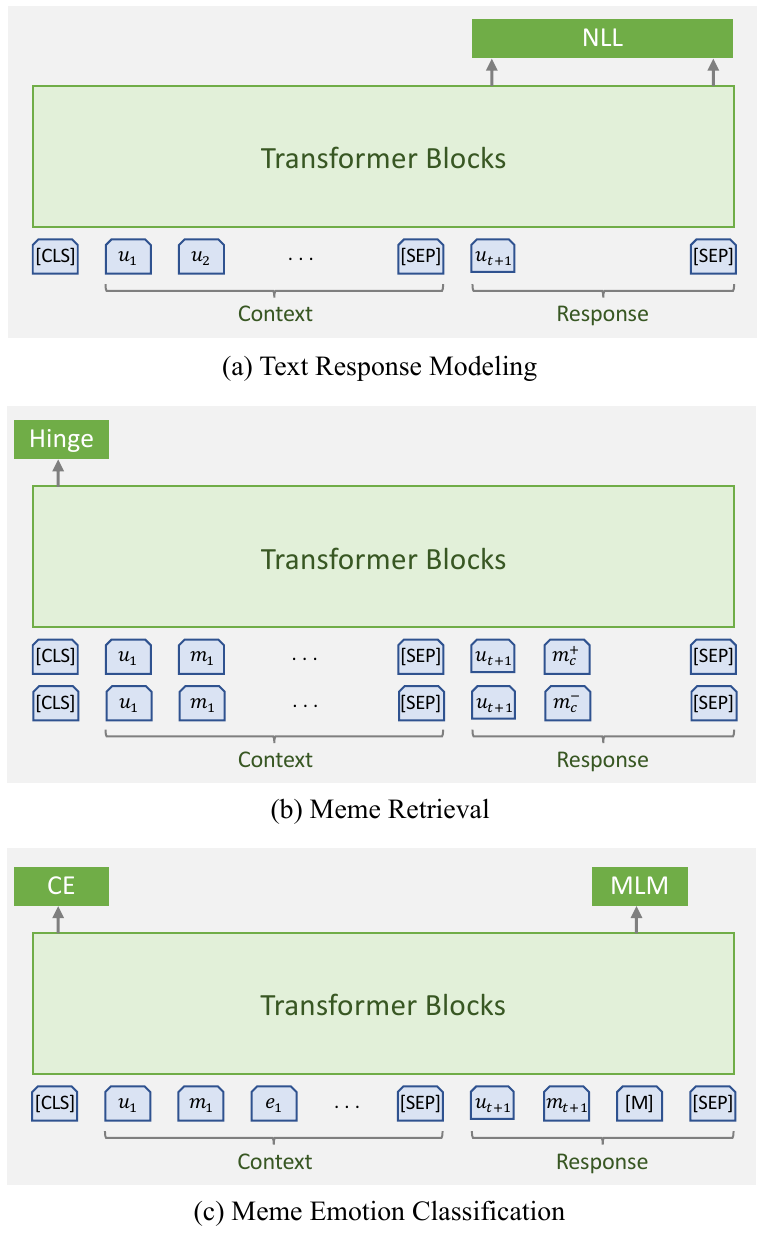}
    \caption{Illustration of network inputs and training objectives for three tasks.}
    \label{fig2:network}
\end{figure}
In this paper, we utilize the pre-trained open-domain dialogue model PLATO-2 \cite{bao2020plato} for text response generation. As illustrated in Figure \ref{fig2:network}(a), the input to the network is the concatenation of context and response. The input representation is calculated as the sum of the token, segment, and position embeddings. The network employs flexible attention mechanisms, where bi-directional attention is enabled for better contextual understanding and uni-directional attention is utilized for auto-regressive response generation. The training objective of text response generation is to minimize the following negative log-likelihood (NLL) loss:
\begin{equation}
	\mathcal{L}_{NLL} = - ~\log p_{\text{generation}}(u_{t+1}|C_t)
\end{equation}

\subsection{Meme Retrieval} 
Meme retrieval is a crucial component in meme incorporated dialogue systems. This task is to select an appropriate meme from the Internet meme set, given the multi-modal dialogue context $C_t$ and text response $u_{t+1}$. 
Formally, we denote Internet meme set as $M = \{m_1, m_2, \cdots, m_n\}$, where $m_i$ is the representation of the $i$-th meme. In this work, we represent memes with textual information, consisting of meme titles and OCR texts (extracted from the memes). Although there exists plenty of vision or multi-modal pre-training works \cite{chen2019uniter,li2020unimo,gan2020large,radford2021learning}, they are less effective at meme feature extraction due to the gap of data distributions between real photos and devised memes. 
Experimental results also suggest that the meme title and OCR text can sufficiently represent the meaning of Internet memes. 

Therefore, we treat the meme retrieval task as a text-matching problem and employ the cross-encoder architecture for relevance estimation. The network overview for meme retrieval is shown in Figure \ref{fig2:network}(b). The input includes the dialogue context $C_t$, the textual response $u_{t+1}$, and one candidate meme $m_c$. During training, a pair of positive and negative samples are fed into the network. The output of [CLS] token is passed through a fully-connected layer, and a following sigmoid function to obtain the relevance probability $p_{\text{matching}}(l_{m_c} = 1|C_t, u_{t+1}, m_c)$, where $l_{m_c}$ stands for the label to choose meme $m_c$ or not given the dialogue context and corresponding textual response. The training objective is to minimize the following margin ranking loss:
\begin{equation}
\begin{split}
\mathcal{L}_{\rm Hinge} = \max(&0, ~p_{\rm matching}(l_{m_c^-} = 1 | C_t, u_{t+1}, m_c^-) \\
 &- p_{\rm matching}(l_{m_c^+} = 1|C_t, u_{t+1}, m_c^+) \\ 
 &+ \alpha)
\raisetag{1.65\baselineskip}
\end{split}
\end{equation}
where $\alpha$ is a pre-defined margin parameter, $m_c^+$ is a positive meme, and $m_c^-$ is a negative meme. 
During training, we enable dynamic random negative sampling, which means that as model training progresses, different sets of negative samples are dynamically sampled in each epoch.

During inference, the Internet meme for the given dialogue context and response is selected as:
\begin{equation}
m^* = \operatorname*{argmax}_{m_c \in M} ~p_{\rm matching}(l_{m_c} = 1|C_t, u_{t+1}, m_c) 
\label{equation:combination}
\end{equation}

\subsection{Meme Emotion Classification} 
\label{Meme Emotion Classification section}
Rather than classify the sentiment of an Internet meme, this task aims to predict the meme emotion type situated in the dialogue context.
Considering that the emotions of two interlocutors seldom encounter abrupt changes (or to some extent the changes might be traceable), we propose to model the emotion flow (EF) in multi-turn conversations. The textual descriptions of emotions are integrated into the dialogue context. Specifically, the utterance at turn $i$ is composed of the utterance text, meme text, and textual emotion description, i.e., $\langle u_i, m_i, e_i\rangle$. The meme emotion recognition can be considered as a classical sequence classification task, and the training objective is to minimize the standard cross-entropy loss. 

Additionally, we introduce an auxiliary task of emotion description prediction (EDP) to boost meme emotion recognition performance.
As shown in Figure \ref{fig2:network}(c), the auxiliary task is to recover the masked tokens (i.e., textual emotion description in the response) by minimizing the masked language model (MLM) loss \cite{devlin2018bert}. In this way, the training objective of meme emotion classification is to minimize the following integrated loss:
\begin{equation}
\mathcal{L} = \mathcal{L}_{CE} + \mathcal{L}_{MLM}
\end{equation}
where $\mathcal{L}_{CE}$ is the cross-entropy loss of the classification task and $\mathcal{L}_{MLM}$ denotes the MLM loss of the emotion description prediction task. 

\section{Experiments}
In the DSTC10 MOD challenge, one open-domain dialogue dataset incorporated with Internet memes is constructed. The memes used in the dataset have been annotated with titles. The MOD test set is divided into easy test version and hard test version. The latter one containing unseen memes is used to evaluate the ability of dialogue systems to exploit new Internet memes. Detailed statistics of the dataset are summarized in Table \ref{tab:dataset_statistic}. 

\subsection{Settings}
The evaluation of the MOD challenge covers the following three tasks:
\begin{itemize}
    \item Task1: Text Response Modeling. Given a dialogue context, the model needs to produce a coherent and informative text response. The automatic evaluation metrics of this task include BLEU-2/4 \cite{papineni2002bleu}, DIST-1/2 \cite{li2015diversity}.
    \item Task2: Meme Retrieval. Given a multi-modal dialogue context and a text response, the model needs to retrieve an appropriate Internet meme. The evaluation metrics of this task include Recall\_10@1, Recall\_10@3, Recall\_10@5, and MAP.
    \item Task3: Meme Emotion Classification. The model needs to predict the corresponding emotion type of the used Internet meme. The evaluation metrics of this task include Accuracy@1, Accuracy@3, and Accuracy@5.
\end{itemize}
\begin{table}
	\centering
	\includegraphics[width=0.48\textwidth]{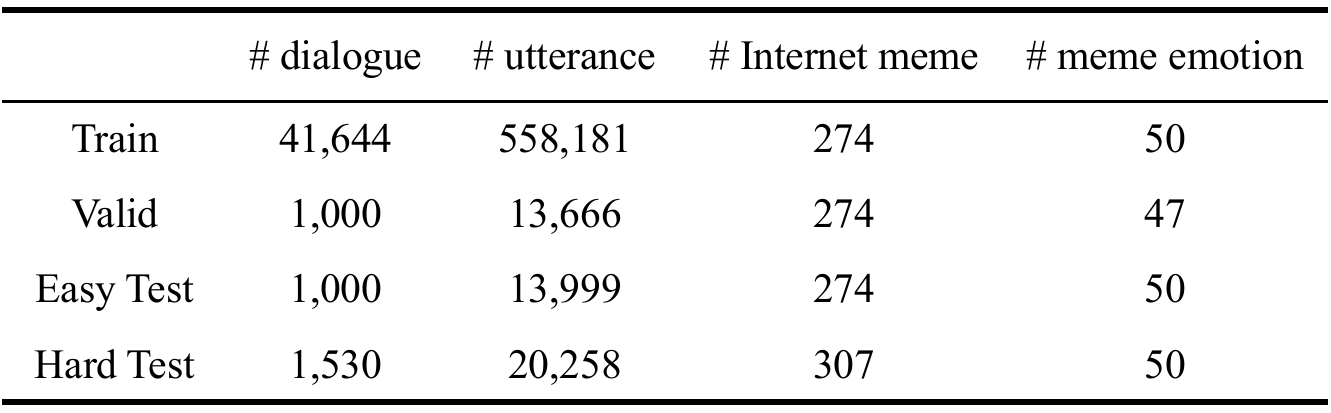}
	\caption{Statistics of the MOD dataset.}
	\label{tab:dataset_statistic}
\end{table}

\noindent \textbf{Implementation Details} In the experiments, we utilize dialogue pre-training models of PLATO-2 \cite{bao2020plato} to improve the performance of all three tasks. The models have 32 transformer blocks and 32 attention heads, with up to 1.6 billion parameters. The generation model of PLATO-2 is used for Task1. The evaluation model of PLATO-2 is employed for the fine-tuning of Task2 and Task3. 

In Task1, responses are generated using beam search, with a beam size of 5. The maximum sequence length for the context and response is set to 256 and 128, respectively. In Task2, we set the margin parameter $\alpha$ to 0.2 and the ratio of positive training samples to negative ones to 1:5. During the fine-tuning of Task2 and Task3, we use Adam \cite{kingma2014adam} optimizer with a learning rate of 2e-5 and warmup steps of 4000. All the models are fine-tuned for five epochs with a batch size of 64. The implementation is based on the PaddlePaddle framework, and the experiments are carried out on 4 Nvidia Tesla V100 GPUs (32G RAM).


\subsection{Experimental Results}
The experimental results on these three tasks are discussed in the following.

\textbf{Text Response Modeling}
The evaluation results of text response modeling on two test versions are summarized in Table \ref{tab:rank_task1}, with the best score written in bold. The final ranking for this task is based on human evaluation, where five metrics are considered: grammatical correctness, informativeness, naturalness, relevance to the dialogue history, and overall feeling based on the above four metrics. The score of each metric ranges from 1 to 5. The higher, the better. The final human score is the average of the above five metric scores. We rank second on the easy version and first on the hard version. 
From Table \ref{tab:rank_task1}, it can be observed that our automatic evaluation results are relatively poor, especially on the metrics of BLEU-2/4, while the human evaluation results are relatively competitive.
This phenomenon further verifies that the correlation between automatic evaluation metrics and human evaluation is weak in open-domain conversations \cite{liu2016not}.
\begin{table}[t]
	\centering
	\includegraphics[width=0.48\textwidth]{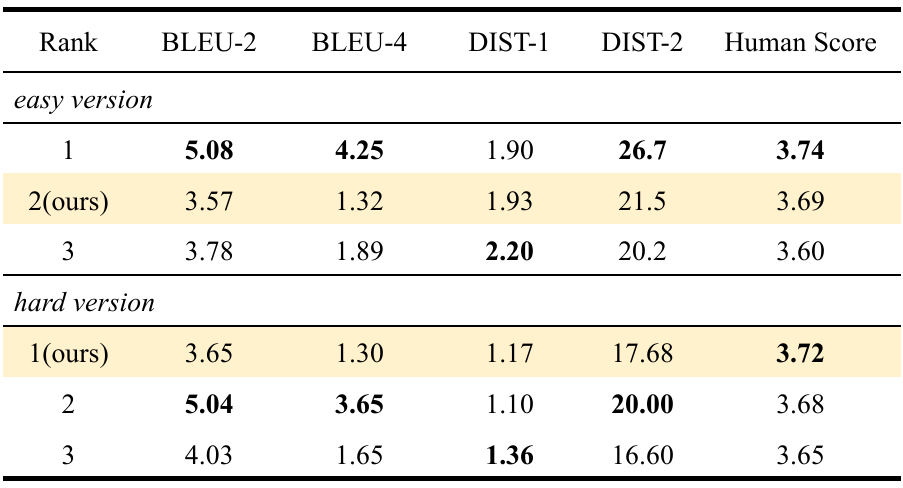}
	\caption{Task1 evaluation results on two test versions, with the best score written in bold.}
	\label{tab:rank_task1}
\end{table}
\begin{table}[t]
	\centering
	\includegraphics[width=0.48\textwidth]{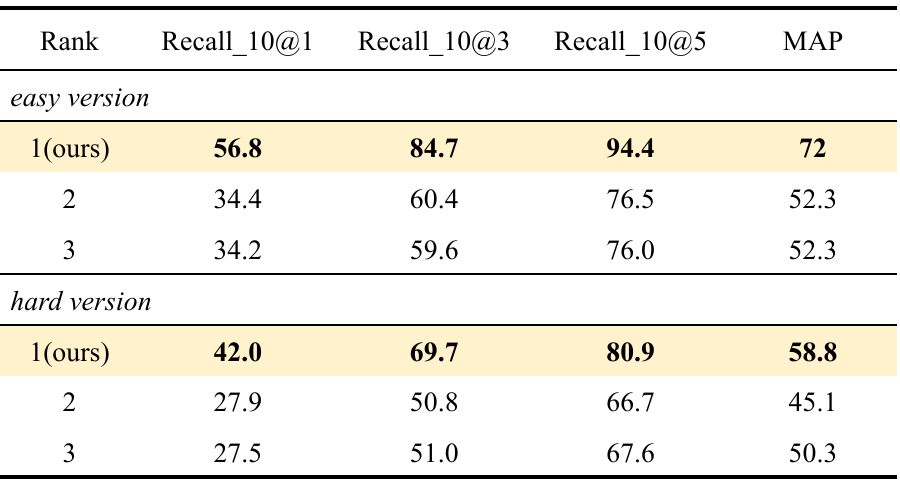}
	\caption{Task2 evaluation results on two test versions, with the best score written in bold.}
	\label{tab:rank_task2}
\end{table}
\begin{table}[h!]
	\centering
	\includegraphics[width=0.48\textwidth]{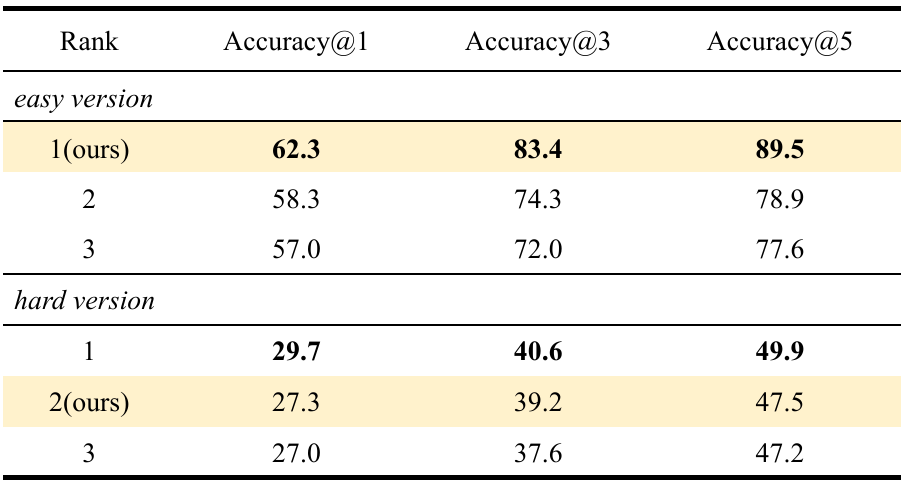}
	\caption{Task3 evaluation results on two test versions, with the best score written in bold.}
	\label{tab:rank_task3}
\end{table}

\textbf{Meme Retrieval}
The evaluation results of meme retrieval on two test versions are summarized in Table \ref{tab:rank_task2}, with the best score written in bold. The final ranking for this task is based on the Recall\_10@1 score, which is the fraction of the ground-truth Internet meme ranked first among ten meme candidates. Our proposed method obtains first place on both versions and outperforms other teams by a large margin. Possible reasons behind such improvements are discussed as follows: 
\begin{itemize}
    \item The modality of dialogue context and memes is unified by representing Internet memes with texts, making it easier for the model to estimate the relevance. Furthermore, with the pre-trained language models, our text-matching strategy has the generalization ability to retrieve the unseen memes that are not available during training.
    \item Different from the dual-encoder architecture \cite{mazare2018training, zang2021photochat}, which performs self-attention over the input and the candidate separately, we employ the cross-encoder architecture to yield rich interactions between the dialogue context and the meme candidate. With this interaction-based text-matching, our model can retrieve appropriate memes more effectively. 
\end{itemize}

\textbf{Meme Emotion Classification}
The evaluation results of meme emotion classification on two test versions are summarized in Table \ref{tab:rank_task3}, with the best score written in bold. The final ranking for this task is based on the Accuracy@1 score, which is the fraction of the ground-truth emotion type that obtains the highest score among all emotion types. We rank first on the easy version and second on the hard version.
In particular, our meme emotion classification model, combining the EF and EDP strategies, achieves 4.0\% absolute improvement on Accuracy@1 on the easy test set over the second-ranked team (62.3\% vs. 58.3\%). On the hard test set, the performance of all the models degenerates significantly, revealing that the emotion classification ability for unseen memes is still weak.

\begin{figure*}[h]
    \centering
    \includegraphics[width=\textwidth]{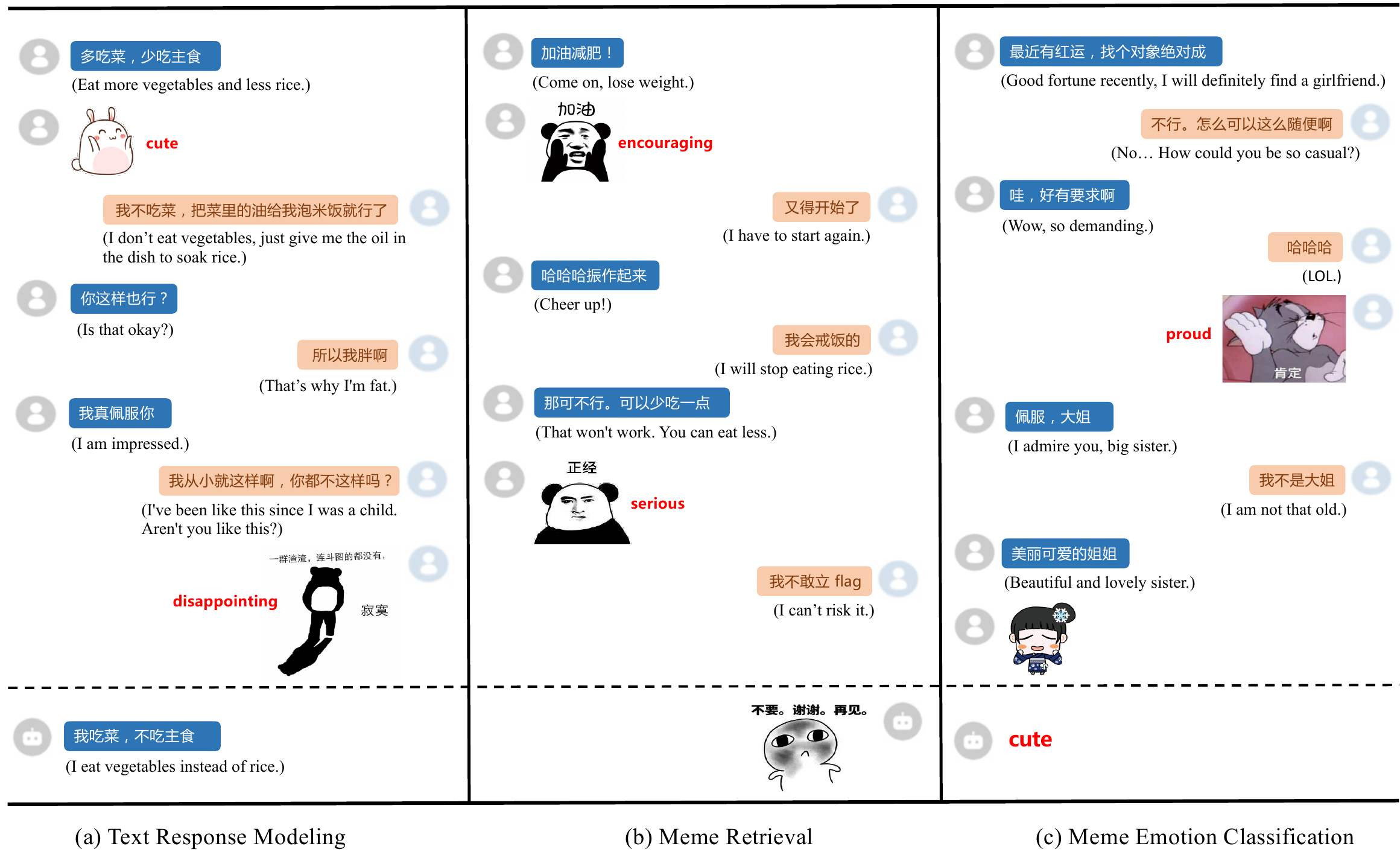}
    \caption{Examples of the input (upper) and our output (bottom) for each task.}
    \label{fig:analysis_case}
\end{figure*}
\subsection{Case Analysis}     
To further analyze the performance of our proposed methods, several examples are provided in Figure \ref{fig:analysis_case}. 
As shown in the left example, our model is able to generate a coherent and informative response.
The interlocutor on the left-hand side seems to be a vegetarian, and our model generates a response consistent with the persona.
In the middle and right examples, our model is able to retrieve a relevant meme and accurately identify the emotion type contained in the meme.
These dialogue examples suggest that our system can generate a natural and informative response incorporated with an appropriate meme.

\subsection{Ablation Study}
In this section, several ablation studies are carried out on the validation set to better understand the contribution of each component.

\subsubsection{Meme Retrieval}
To evaluate the generalization ability of our matching model, we select 20 memes to formulate the unseen validation set and remove corresponding samples from the train set. 
The experimental results on the unseen validation set by the matching models trained on the original train set and the filtered train set are summarized in Table \ref{tab:ablation_task2_2}. The results indicate that our model has the generalization ability to exploit unseen memes. 
However, the gap between original and filtered (50.4\% vs. 43.6\% on Recall\_10@1) suggests that there is still some room for improvement on the model's generalization ability.

\begin{table}[t]
	\centering
	\includegraphics[width=0.48\textwidth]{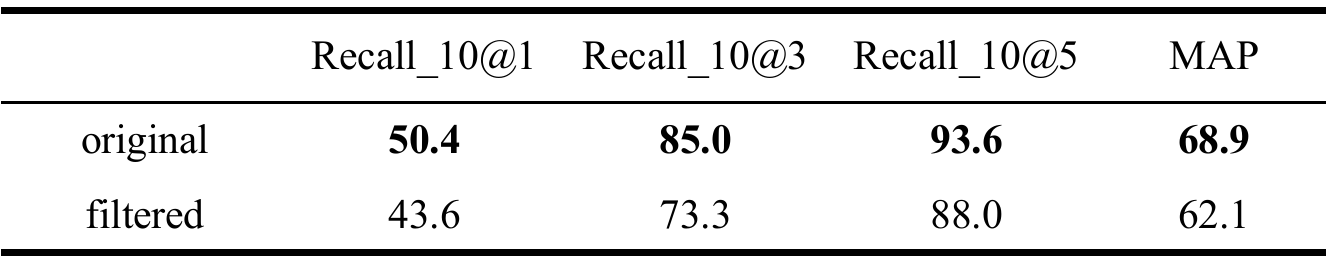}
	\caption{Task2 ablation study on the unseen validation set.}
	\label{tab:ablation_task2_2}
\end{table}
\begin{table}[h!]
	\centering
	\includegraphics[width=0.48\textwidth]{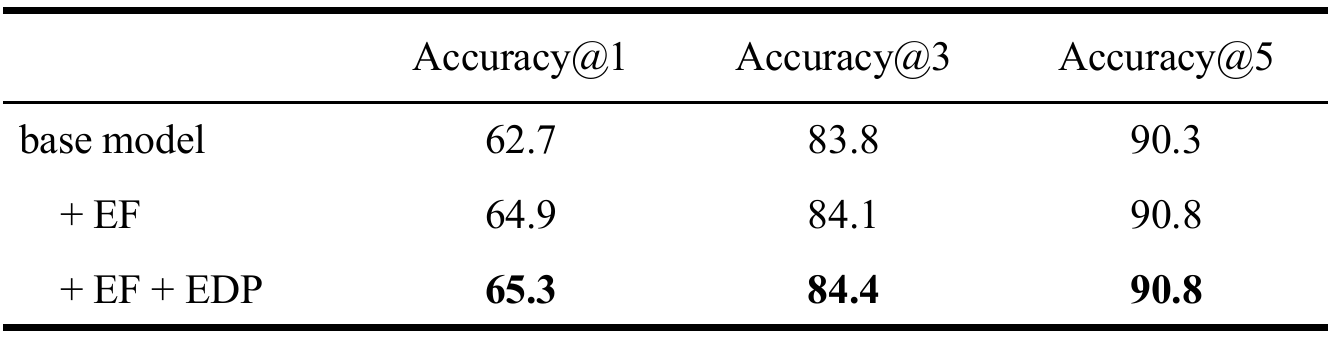}
	\caption{Task3 ablation study on the validation set. EF refers to the modeling of emotion flow. EDP refers to the task of emotion description prediction.}
	\label{tab:ablation_task1}
\end{table}
\subsubsection{Meme Emotion Classification}
To boost the performance of meme emotion classification, we propose to model the emotion flow (EF) in multi-turn conversations and introduce an auxiliary task of the emotion description prediction (EDP). To analyze the effects of these two components, we carry out the ablation studies, and evaluation results are summarized in Table \ref{tab:ablation_task1}. Compared to the base model, the incorporation of EF modeling gives rise to a significant improvement (+2.2\% on Accuracy@1). The combination of EF and EDP obtains +2.6\% absolute improvement on Accuracy@1 (65.3\% vs. 62.7\%), verifying the effectiveness of our proposed strategies.

\section{Related Work}
In this section, we will discuss related works on multi-modal conversation and emotion recognition.

There are several works that attempt to incorporate multi-modal information into conversations. \citeauthor{das2017visual} introduces the task of VisDial, where the AI agent needs to hold a meaningful conversation with humans and answer questions about the contents of the input image. In addition to the conversational question answering, there are some other tasks where natural and engaging conversations are conducted based on a shared image, such as image-grounded conversations \cite{mostafazadeh2017image}, and image-chat \cite{shuster2018image}. Recently, the PhotoChat dataset \cite{zang2021photochat} is presented, which focuses on the photo-sharing behavior in online messaging and aims to improve the photo-sharing experience in conversations. Unlike the above works concentrated on photos, the MOD dataset incorporates Internet memes into open-domain conversations to enhance communication expressiveness.


In open-domain conversation, it is crucial to recognize the emotional state accurately and generate a response appropriately \cite{rashkin2018towards}. To boost emotion detection in conversations, HiTrans \cite{li2020hitrans} utilizes BERT \cite{devlin2018bert} as the low-level transformer to generate utterance representations and employs another high-level transformer to obtain context representations. In TUCORE-GCN \cite{lee2021graph}, the task of emotion recognition is treated as a dialogue-based relation extraction, where a dialogue graph is constructed and a graph convolution network is employed for relation classification. In addition to the text-based conversations, emotion has been widely analyzed in some areas of computer vision, such as facial expression recognition \cite{minaee2021deep}, image emotion classification \cite{yang2017joint}, and so on. In this work, rather than classify the sentiment of an Internet meme, this task aims to predict the meme emotion type situated in the dialogue context.


\section{Conclusion}
In this paper, we introduce our solutions for the DSTC10 MOD challenge.
Firstly, we leverage a large-scale pre-trained dialogue model for coherent and informative response generation. Secondly, based on interaction-based text-matching, our approach can retrieve appropriate memes with good generalization ability.
Thirdly, we propose to model the emotion flow (EF) in conversations and introduce an auxiliary task of emotion description prediction (EDP) to boost the performance of meme emotion recognition.
Comprehensive experiments have been conducted on the MOD dataset. Experimental results demonstrate that our methods can effectively incorporate Internet memes into dialogue systems and accurately recognize the meme emotion. Our methods with competitive performance achieve first place in four out of six leaderboards and second place in the others.

\section*{Acknowledgments}
We would like to thank the anonymous reviewers for their constructive suggestions; Xinxian Huang for the helpful discussions.



\bibliography{aaai22}

\end{document}